\documentclass[runningheads]{llncs}
\usepackage[T1]{fontenc}
\usepackage{{booktabs}}
\usepackage{graphicx}
\usepackage{hyperref}
\usepackage{color}
\usepackage{xcolor}
\definecolor{grey}{RGB}{77,77,77}
\definecolor{myred}{RGB}{234,107,102}

\usepackage{tikz}
\newcommand\circled[2][]{%
  \tikz[baseline=(X.base)] 
    \node (X) [shape=circle, inner sep=0, fill=grey, text=white, #1] {\strut \bfseries #2};%
}

\begin{document}
\title{SkillGPT: a RESTful API service for skill extraction and standardization using a Large Language Model
}

\titlerunning{SkillGPT}
\author{Nan Li
\and
Bo Kang
\and
Tijl De Bie
}
\authorrunning{N. Li et al.}
\institute{IDLAB - Department of Electronics and Information Systems (ELIS), Ghent University, Ghent 9000, Belgium
\email{\{firstname.lastname\}@ugent.be}}

\maketitle              %
\begin{abstract}
We present SkillGPT, a tool for skill extraction and standardization (SES) from free-style job descriptions and user profiles with an open-source Large Language Model (LLM) as backbone. Most previous methods for similar tasks either need supervision or rely on heavy data-preprocessing and feature engineering. Directly prompting the latest conversational LLM for standard skills, however, is slow, costly and inaccurate. In contrast, SkillGPT utilizes a LLM to perform its tasks in steps via summarization and vector similarity search, to balance speed with precision. The backbone LLM of SkillGPT is based on Llama, free for academic use and thus useful for exploratory research and prototype development. Hence, our cost-free SkillGPT gives users the convenience of conversational SES, efficiently and reliably.

\keywords{Skill extraction \and Skill standardization \and Information retrieval \and Large language model.}
\end{abstract}
\section{Introduction}\label{sec:intro}
Automatic \emph{skill extraction} aims to detect skills and qualities from unstructured job descriptions and user profiles without human involvement. \emph{Skill standardization} further maps the free-styled skill annotations onto a standard taxonomy, such as the ``European Skills, Competences, Qualifications and Occupations'' (ESCO) standard \cite{esco}. Skill extraction and standardization (SES) benefit various downstream tasks, such as job recommendation and career path planning.   
The state of art SES methods are mostly supervised \cite{sayfullina2018learning,tamburri2020dataops,green2022development,bhola2020retrieving}, relying on the costly and cumbersome process of human annotation. Weakly supervised techniques proposed such as by Zhang et al. \cite{zhang2022skill} require heavy data preprocessing and feature engineering. BERT-based Large Language Models (LLMs) \cite{devlin2018bert,liu2019roberta} were utilized recently \cite{chernova2020occupational,zhang2022kompetencer,gnehm2022evaluation}, but merely as a refinement of the former static word embeddings.

Recent LLMs have created the possibility of solving SES without human annotation, heavy data-preprocessing or feature engineering. The latest generation (e.g. chatGPT) further shows the power and convenience of conversational interfaces, making code-less interactions possible for domain experts. However, LLMs have some limitations that make them hard to apply directly for SES. Directly prompting a job posting for ESCO codes might well be answered with non-existent codes due to the \emph{hallucination} problem \cite{dziri2022origin,ji2023survey}. To improve the accuracy, ESCO documents should be fed into the LLM as context. Yet, handling \emph{lengthy prompts} of hundreds of pages is not straightforward. Even if the ESCO documents were somehow prompted (e.g. by chunk), model predictions would be \emph{inefficient}, making it infeasible for online usage. Using powerful but \emph{costly and close-sourced} models from OpenAI such as ChatGPT would hinder exploration and evaluation of possible solutions.

These considerations motivated us to design an SES tool to utilize the power of LLMs without sacrificing \emph{accuracy, speed and cost-efficiency}. We call our solution \textbf{SkillGPT}: an API service for skill extraction and standardization from free-style text, that uses \emph{fast} vector similarity search against precomputed ESCO embeddings, retrieves \emph{with high precision} the matching codes given any prompted job description or user profile, and \emph{costs little} (at least for academia) by using the  latest open-sourced LLM Vicuna-13B \cite{peng2023instruction}\footnote{Vicuna is based on Llama \cite{touvron2023llama}, reported to perform comparably to GPT-3, and fine-tuned using data from shareGPT for better alignment with human preferences. Our experiments showed that Vicuna-13B (the number of parameters is 13 billion) has sufficient capacity for SES. More importantly, as it is open-source we can to deploy our model on a local machine with minimal cost. Nonetheless, developers can replace Vicuna with any LLM they want, including paid ones.} (Sec.~\ref{sec:design}). SkillGPT has more merits: it is \emph{multi-faceted} as it can extract multiple ESCO concept types, and it offers \emph{multi-lingual} support for English, French and Dutch (Sec.~\ref{sec:use-cases}). Our code is publicly available at \url{https://github.com/aida-ugent/SkillGPT}.

\section{API Design}\label{sec:design}
\begin{figure}
\center
\includegraphics[width=0.9\textwidth]{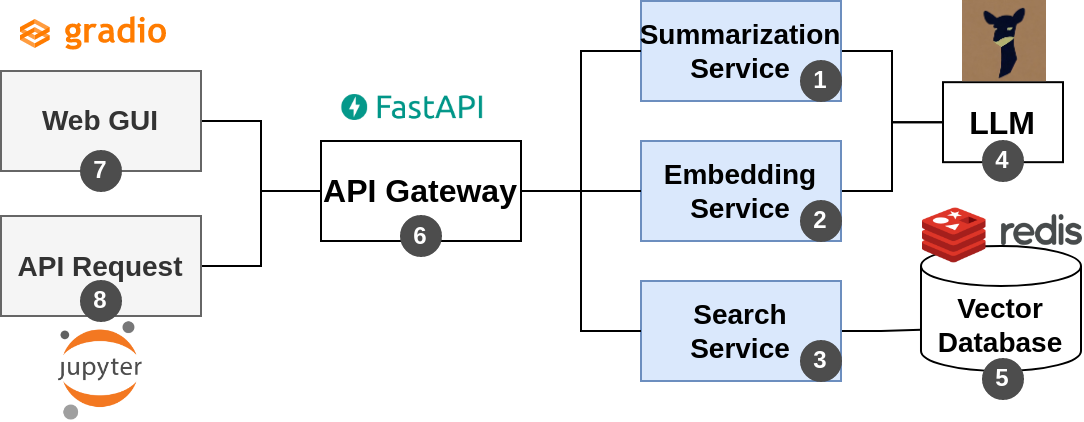}
\caption{Main components of SkillGPT. The specific tools and models are our careful choices for the current version, but our framework is flexible to accomodate other options. Note that this component diagram merely shows the design components, not the data flow.} \label{fig:components}
\end{figure}

\textbf{Fig.~\ref{fig:components}} is a high-level overview of SkillGPT's components, with annotations of the specific models and tools\footnote{We choose the tools through careful considerations and experiments. Detailed justification of our design choices as well as ablation study will be presented in an extended manuscript.} used in our demo.

\begin{figure}
\includegraphics[width=\textwidth]{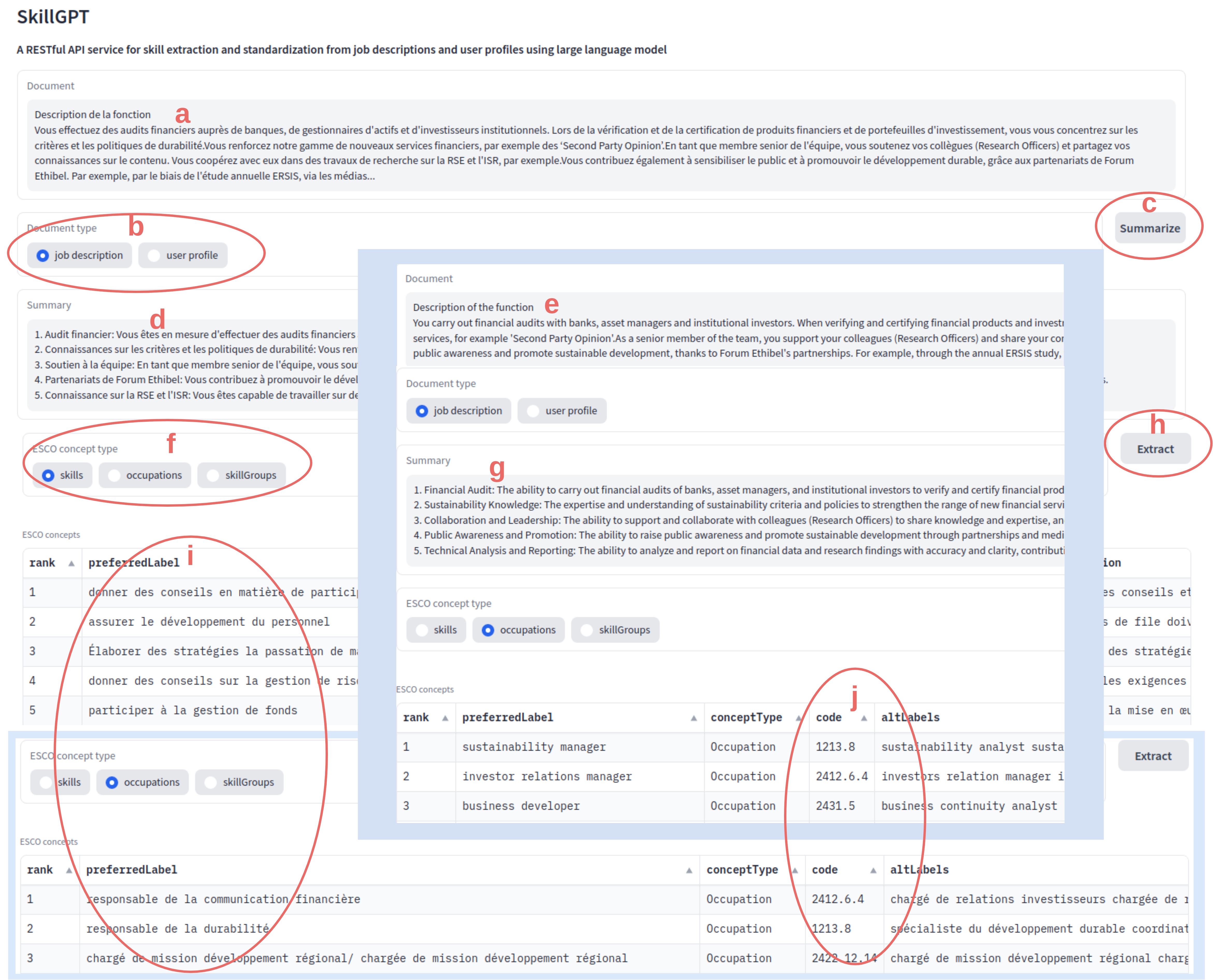}
\caption{SkillGPT usage: multi-faceted and multi-lingual. The screenshots with blue background are from different sessions.} \label{fig:use}
\end{figure}

There are two major phases of SkillGPT's lifecycle. \emph{First} is the system initialization for embedding and storing the ESCO taxonomy, where the ESCO entries are downloaded and converted into structured documents, then their vectorized representations are computed by the backbone LLM\circled[scale=0.8]{4} via a document embedding service\circled[scale=0.8]{2}, and finally stored in the vector database\circled[scale=0.8]{5}.

\emph{Second} is the main functionality of skill extraction and standardization. Extracting free-style textual skills from a job description or user resume is achieved by carefully prompting the LLM for summarization\circled[scale=0.8]{1}. The embedding (via \circled[scale=0.8]{2}) of the distilled text is used as a query to retrieve the top-k similar ESCO terms via vector similarity search\circled[scale=0.8]{3}. 
All these are coordinated by an API gateway\circled[scale=0.8]{6}. 

To perform an one-stop SES or to interact with any individual service, we provide two options: direct interaction with a RESTful API\circled[scale=0.8]{8}, or via graphical user interface\circled[scale=0.8]{7} (e.g. Gradio App).

\section{Use cases}\label{sec:use-cases}
The current version SkillGPT supports 2 document types (job description/user resume), 3 ESCO concept types (Skill/Occupation/Occupation group) and 3 languages (En/Fr/Nl), yielding \textbf{18} possible use cases. Due to space limitation, we only show the following condensed exemplary use cases, and kindly refer to our demo videos  (\url{http://bokang.io/videos/SkillGPT.mp4}) for more.
\emph{First}, a user pastes a free-styled document in French (Fig.~\ref{fig:use}\textcolor{myred}{\bfseries{a}}), English(Fig.~\ref{fig:use}\textcolor{myred}{\bfseries{e}}), or Dutch into the top textbox, followed by choosing the type of the document between job description and user profile (Fig.~\ref{fig:use}\textcolor{myred}{\bfseries{b}}). \emph{Then} the user clicks \texttt{Summarize} (Fig.~\ref{fig:use}\textcolor{myred}{\bfseries{c}}) to let the backbone LLM distill the skills contained in the document. The output skill list will be in the same language of the original input (Fig.~\ref{fig:use}\textcolor{myred}{\bfseries{d}} and \textcolor{myred}{\bfseries{g}}). \emph{Next}, the user can choose which ESCO concept type (Fig.~\ref{fig:use}\textcolor{myred}{\bfseries{f}}) to standardize the free-style skill descriptions, and the corresponding most plausible ESCO terminologies will be returned (Fig.~\ref{fig:use}\textcolor{myred}{\bfseries{i}}). One  \emph{limitation} is that the extracted codes for the same content in different languages, although they mostly overlap, may differ somewhat (Fig.~\ref{fig:use}\textcolor{myred}{\bfseries{j}}). Our test trials also showed that SkillGPT seems to perform better in English documents than Dutch or French. The LLM probably models the English language better than other languages due to the non-uniform volume distribution of the training corpus. There is also some randomness in the generation procedure of skill summaries.

\section{Conclusion}\label{sec:conclusion}
We presented SkillGPT, a flexible and easy-to-use tool for skill extraction and standardization from free-styled job descriptions and user resumes. SkillGPT is \emph{efficient}, \emph{economical} and delivers often truthful and \emph{plausible} results, thus meeting our design goals.
However, SkillGPT in its current state has some \emph{limitations}. First, treating the summarized text as a single document might cause certain subtle skills to be lost, since the synthesized embedding might be dominated by the most salient qualities. Second, many options for optimizing the performance of LLMs, such as fine-tuning and prompt engineering, have not properly been examined, due to time limitations and, more importantly, that how to harness LLMs is a booming field where new hacks and tricks are proposed everyday. 
In \emph{future}, we plan to address the aforementioned limitations, perform both qualitative and quantitative evaluations on SES and various downstream tasks in e-recruitment recommendation \cite{mashayekhi2022challengebased}, and optimize for smaller languages as well as support the full range of 25 European Union languages. 

\subsubsection{Acknowledgements} The research leading to these results has received funding from the European Research Council under the European Union's Seventh Framework Programme (FP7/2007-2013) (ERC Grant Agreement no. 615517), and under the European Union’s Horizon 2020 research and innovation programme (ERC Grant Agreement no. 963924), from the Special Research Fund (BOF) of Ghent University (BOF20/IBF/117), from the Flemish Government under the ``Onderzoeksprogramma Artificiële Intelligentie (AI) Vlaanderen'' programme, and from the FWO (project no. G0F9816N, 3G042220).

\bibliographystyle{splncs04}
\bibliography{ref}
\end{document}